%% file: acl_latex.tex
\documentclass[11pt]{article}

\usepackage{acl}

\usepackage{times}
\usepackage{latexsym}
\usepackage{amsmath}
\usepackage{amssymb}
\usepackage{mathtools}
\usepackage{graphicx}
\newcommand{\ours}{\textsc{DART}\thinspace}
\usepackage[T1]{fontenc}
\usepackage[most]{tcolorbox}
\usepackage{bm}
\usepackage[utf8]{inputenc}
\usepackage{multirow}  
\usepackage{graphicx}  
\usepackage{microtype}
\usepackage{colortbl}
\usepackage{inconsolata}
\usepackage{algorithm}
\usepackage{algpseudocode}
\usepackage{graphicx}
\usepackage{tabularx}
\usepackage{hyperref}
\usepackage{url}
\usepackage{multicol}
\usepackage{aliascnt}
\usepackage{adjustbox}
\usepackage{siunitx} 
\usepackage{booktabs}
\usepackage{multirow} 
\usepackage{enumitem}
\usepackage{booktabs} 
\usepackage{amssymb}  
\usepackage{amsmath}  
\usepackage{graphicx} 
\usepackage{longtable} 
\usepackage{graphicx}
\usepackage{subcaption}
\usepackage{calc}
\usepackage[most]{tcolorbox}
\usepackage{listings}
\definecolor{uclablue}{rgb}{0.15, 0.45, 0.68}
\definecolor{uclablue}{rgb}{0.15, 0.45, 0.68}
\definecolor{mypink}{RGB}{255, 192, 203}
\hypersetup{
    breaklinks,
    citecolor=uclablue,
    colorlinks=true,
}
\newtcolorbox{AIbox}[2][]{aibox,title=#2,#1}
\tcbset{
  aibox/.style={
    top=10pt,
    colframe=black,
    colbacktitle=black,
    coltitle=white,
    center,
  }
}

\lstdefinelanguage{prompt}{
    basicstyle=\scriptsize\ttfamily, 
    mathescape=true,        
    escapebegin=\color{latentcolor},  
    escapeend={},
    escapechar=@,
    stringstyle = \color{myorange},
    showstringspaces = false,
    moredelim = [s][\color{mypink}]{`}{`},
    moredelim = [s][\color{mybrown}]{```json}{```},
    moredelim = [s][\color{latentcolor}]{<StartOfLatent>}{<EndOfLatent>},
    literate = %
        {\ \ a.\ }{{\textcolor{mypurple}{\ \ a.\ }}}5
        {\ \ b.\ }{{\textcolor{mypurple}{\ \ b.\ }}}5
        {\ \ c.\ }{{\textcolor{mypurple}{\ \ c.\ }}}5
        {\ \ d.\ }{{\textcolor{mypurple}{\ \ d.\ }}}5
        {\ \ e.\ }{{\textcolor{mypurple}{\ \ e.\ }}}5
        {\ \ f.\ }{{\textcolor{mypurple}{\ \ f.\ }}}5
        {\ \ g.\ }{{\textcolor{mypurple}{\ \ g.\ }}}5
        {\ \ h.\ }{{\textcolor{mypurple}{\ \ h.\ }}}5
        {\ I.\ }{{\textcolor{mypurple}{\ I.\ }}}4
        {\ II.\ }{{\textcolor{mypurple}{\ II.\ }}}5
        {\ III.\ }{{\textcolor{mypurple}{\ III.\ }}}6
        {\ IV.\ }{{\textcolor{mypurple}{\ IV.\ }}}5
        {\ V.\ }{{\textcolor{mypurple}{\ V.\ }}}4
}

\newtcblisting[auto counter, number within=subsection]{prompt}[4][]{%
  before skip=6pt, after skip=6pt,    
  top=6pt, bottom=6pt, left=7pt, right=7pt, 
  enhanced, breakable,
  colback=white, colframe=gray!65,
  boxrule=0.6pt, arc=2pt,
  drop shadow={black!15},             
  listing only,
  listing options={
    language=prompt,
    upquote=true,
    basicstyle=\scriptsize\ttfamily \setlength{\baselineskip}{1.1\baselineskip},
    columns=fullflexible,
    breaklines=true,
    breakindent=0pt,
    xleftmargin=\ifx\empty#2\empty0pt\else#2\fi,
    xrightmargin=\ifx\empty#3\empty0pt\else#3\fi,
    aboveskip=0pt,                    
    belowskip=0pt,                    
    showstringspaces=false,
  },
  title={\ifstrempty{#4}{Prompt \thetcbcounter}{Prompt \thetcbcounter: #4}},
  colbacktitle=gray!6, coltitle=black,
  attach boxed title to top left={yshift=-1mm, xshift=6pt},
  boxed title style={colback=gray!6, boxrule=0pt, sharp corners},
  #1
}

\newtcblisting[auto counter, number within=subsection]{showcase}[4][]{%
  before=\par\vspace{\baselineskip},
  after=\par,
  width=\linewidth,
  enhanced,
  arc=0em,
  boxrule=1pt,
  listing only,
  listing options={
    language=prompt,
    upquote=true,
    basicstyle=\scriptsize\ttfamily \setlength{\baselineskip}{1.1\baselineskip},
    breaklines=true,
    breakindent=0pt,
    xleftmargin=\ifx\empty#2\empty-12pt\else#2\fi,
    xrightmargin=\ifx\empty#3\empty-5pt\else#3\fi,
    aboveskip=-4pt,
    belowskip=-4pt,
    columns=fullflexible,
    },
  colback=white,
  colframe=gray,
  colbacktitle=gray!5,
  coltitle=black,
  attach boxed title to top center={yshift=-3mm},
  box align=center,
  parbox=false,
  title={\ifstrempty{#4}{Example \thetcbcounter}{Example \thetcbcounter: #4}},
  #1
}

\definecolor{linkColor}{rgb}{0.2,0.4,0.6}
\definecolor{myblue}{HTML}{0379AC}
\definecolor{myred}{HTML}{A50E50}
\definecolor{myorange}{RGB}{238, 133, 74}
\definecolor{latentcolor}{named}{cyan}
\definecolor{normalcolor}{RGB}{0, 0, 0}
\usepackage{marvosym}

\newtcolorbox{llmcontainer}{
    enhanced,
    frame hidden,
    borderline north={0.8pt}{0pt}{black},
    borderline south={0.8pt}{0pt}{black},
    colback=white,
    left=0pt,
    right=0pt,
    breakable,
}

\title{Discovery and Reinforcement of Tool-Integrated \\Reasoning Chains via Rollout Trees}

\author{
  Kun Li$^{1,2}$\thanks{$\;\;$Equal contributions.}, Zenan Xu$^{2}$\footnotemark[1], Junan Li$^{1}$, Zengrui Jin$^{3}$, \\ \bf Jinghao Deng$^{4}$, Zexuan Qiu$^{1,2}$, Bo Zhou$^2$$\thanks{$\;\;$Corresponding author.}$\\
  $^{1}$The Chinese University of Hong Kong \quad $^{2}$LLM Department, Tencent\hspace{0.3cm} \\
  $^{3}$Tsinghua University \quad $^{4}$Tencent\hspace{0.5cm} \\
  \texttt{chaysezhou@tencent.com}
  }

\begin{document}
\maketitle
\thispagestyle{firstpage}
\begin{abstract}
Tool-Integrated Reasoning has emerged as a key paradigm to augment Large Language Models (LLMs) with computational capabilities, yet integrating tool-use into long Chain-of-Thought (long CoT) remains underexplored, largely due to the scarcity of training data and the challenge of integrating tool-use without compromising the model's intrinsic long-chain reasoning. In this paper, we introduce  \textbf{DART} (\textbf{D}iscovery \textbf{A}nd \textbf{R}einforcement of \textbf{T}ool-Integrated Reasoning Chains via Rollout Trees), a reinforcement learning framework that enables spontaneous tool-use during long CoT reasoning without human annotation. DART operates by constructing dynamic rollout trees during training to \textit{discover} valid tool-use opportunities, branching out at promising positions to explore diverse tool-integrated trajectories. Subsequently, a tree-based process advantage estimation identifies and credits specific sub-trajectories where tool invocation positively contributes to the solution, effectively \textit{reinforcing} these beneficial behaviors. Extensive experiments on challenging benchmarks like AIME and GPQA-Diamond demonstrate that DART significantly outperforms existing methods, successfully harmonizing tool execution with long CoT reasoning.

\end{abstract}
\section{Introduction}




Tool-Integrated Reasoning (TIR) \cite{schick2023toolformer, gou2024toratoolintegratedreasoningagent, chen-etal-2025-acebench} has emerged as a promising paradigm to augment large language models (LLMs) with computational capabilities, addressing inherent limitations in precise calculation, symbolic manipulation, and complex equation solving that hinder pure natural language reasoning approaches. 

However, existing TIR efforts have predominantly focused on short Chain-of Thought (CoT) \cite{wei2022chain}, largely due to the modest level of implementation complexity---they can be implemented via prompting \cite{chen2023programthoughtspromptingdisentangling,zhang-etal-2024-natural}, supervised finetuning (SFT) with annotated data \cite{yang2024qwen25mathtechnicalreportmathematical,feng2025retool}, or direct reinforcement learning (RL) \cite{li2025torl, xue2025simpletirendtoendreinforcementlearning}. 
Long CoT by Large Reasoning Models (LRMs) has been demonstrated as a remarkable capability in complex reasoning \cite{openai2024openaio1card, deepseekai2025deepseekr1incentivizingreasoningcapability}. On the other hand, it still  heavily relies on natural language as the reasoning medium, and the integration of tool-use into long CoT has been underexplored. This limitation stems primarily from the design of LRM post‑training for reasoning: the SFT employs curated data consisting of pure-text long CoT, and the subsequent RL rewards the models based on the correctness of final answers with little attention to intermediate reasoning behaviors \cite{deepseekai2025deepseekr1incentivizingreasoningcapability, yang2025qwen3technicalreport}.

It is infeasible to apply the aforementioned TIR methods, originally devised for short CoT, directly to long CoT, for the following reasons:
(1) Prompting with tool-use instructions affects only the answer part instead of the thinking part (long CoT) 
in LRMs' responses (seen in \S~\ref{sec:main_results}), hindering  the applicability of prompting-based and even direct RL-based methods (due to the low hit rate in the rollout phase).
(2) For supervised finetuning, data annotation incurs non-negligible cost, especially for long CoT. Furthermore, SFT on new data may perturb the original distribution of LRMs, potentially impairing their long CoT capability.

Recent LLMs (or LRMs) already possess non-trivial tool-use proficiency thanks to pre-training on extensive code- and tool-related data \cite{pmlr-v202-gao23f,chen2023programthoughtspromptingdisentangling, gou2024toratoolintegratedreasoningagent, wang2023mathcoderseamlesscodeintegration, yang2024qwen25mathtechnicalreportmathematical}. Nevertheless, such capability plays no role during long CoT reasoning. This arises primarily because tool‑use behaviors and long CoT reasoning are learned from separate data sources during post‑training, resulting in two disjoint output distributions.

Based on all the considerations above, in this paper we explore integrating tool-use behaviors into long CoT without annotated data. We enable spontaneous tool-use in long CoT by fusing the output distributions of Tool-Integrated-Reasoning and long CoT, leveraging a self‑bootstrap procedure rather than training on newly curated data. This design offers twofold advantages: first, \textit{it eliminates the cost  associated with cold-start data curation}. Second, \textit{it preserves and builds on the model’s intrinsic long CoT capabilities by aligning tool-use with, rather than supplanting, its native reasoning patterns.} 

Specifically, 
at the core of our approach is a dynamic rollout mechanism in our reinforcement learning process, which elicits and reinforces tool-integrated reasoning trajectories. During the rollout phase, we construct a rollout tree by injecting tool-related hints at the high-entropy yet contextually appropriate positions, effectively merging tool invocation into long-chain reasoning. Furthermore, we introduce a tree-based process advantage estimation. This fine-grained signal enables the model to distinguish and reinforce the specific sub-trajectories where tool invocation genuinely contributes to solving the problem.
We validate our approach on three challenging benchmarks. Experimental results demonstrate that our approach significantly outperforms existing methods, and successfully preserves the long CoT ability of LRMs while organically integrating tool-use.

In summary, our contributions are as follows: (\textit{\romannumeral 1\relax}) We introduce \textsc{DART}, a novel RL framework that integrates tool-use into long Chain-of-Thought reasoning without annotated data, eliminating annotation costs and preserving the model’s original reasoning capabilities. (\textit{\romannumeral 2\relax})  We propose a dynamic tree rollout mechanism to discover tool-use opportunities and tree-based advantage estimation for precise process supervision, effectively reinforcing beneficial tool-integrated behaviors. (\textit{\romannumeral 3\relax})  Extensive experiments show that \textsc{DART} achieves superior performance on complex reasoning tasks.

\input{sections/method}
\input{sections/experiments}
\input{sections/related_works}
\section{Conclusion}
In this paper, we propose \textsc{DART}, a reinforcement learning framework that trains Large Reasoning Models to integrate tool-use into long Chain-of-Thought reasoning without relying on any annotated data. By leveraging the proposed rollout tree mechanism, our approach enables the model to discover and reinforce beneficial tool-integrated trajectories. This method successfully preserves the model's native reasoning capabilities while enhancing tool-use capabilities, achieving superior performance across complex reasoning tasks.

\section*{Limitations}
In this work, we primarily focus on integrating Python interpreters for mathematical and scientific reasoning tasks (e.g., AIME, GPQA). While this serves as a robust testbed for complex logic, the potential of \textsc{DART} in other domains (such as fact-checking or creative writing) and with other types of tools (such as search engines or knowledge graph APIs) has yet to be fully explored. We believe extending our framework to support a wider range of tools is a promising direction for future research.

Although the proposed rollout‑tree construction effectively bridges the gap between long chain‑of‑thought reasoning and tool-use, its iterative nature inevitably introduces additional latency during the training phase when compared to reinforcement learning with standard rollout or supervised fine‑tuning. In this work, we have implemented asynchronous rollout with the SGLang inference engine, which significantly alleviates the latency. Future work could explore more efficient tree construction strategies or asynchronous mechanisms to further optimize the training throughput.

\section*{Ethics Statement}
A central component of our proposed framework, \textsc{DART}, involves the autonomous generation and execution of Python code by Large Language Models. While this significantly enhances reasoning capabilities, it introduces potential risks associated with executing arbitrary code. In our experiments, all code execution was strictly confined within a stateless, non-interactive, and isolated sandbox environment (SandboxFusion) to prevent any unauthorized access to the host system or external networks. We strongly emphasize that any real-world deployment of similar tool-integrated reasoning models must implement rigorous sandboxing and security protocols to mitigate the risks of malicious code execution or unintended side effects.

\bibliography{custom}

\appendix

\input{sections/appendix}

\end{document}

%% file: sections/method.tex
\section{Methodology}
\begin{figure*}[ht]
\centering
\includegraphics[width=1\textwidth]{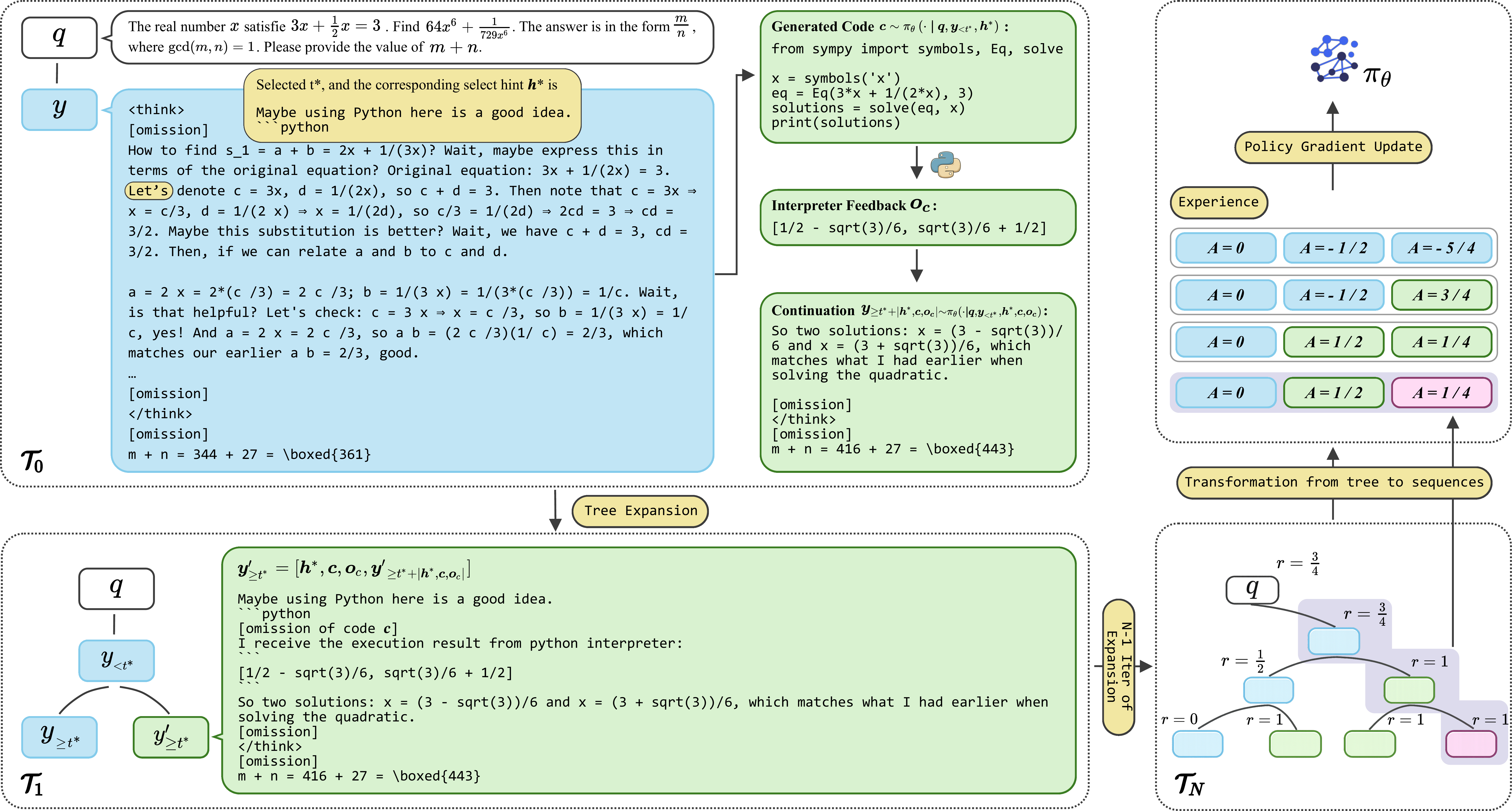}
\caption{Overview of our approach on a running example with $M=1$, $N=3$. 
$\underline{\mathbf{\mathbf{\mathcal{T}_0\rightarrow\mathcal{T}_1\rightarrow...\rightarrow\mathcal{T}_N}}}$: At each expansion step, a promising $(t^*, \bm{h}^*)$ is selected, where $t^*$ is an intermediate position within the existing trajectories (e.g., $\bm{y}$ in $\mathcal{T}_0$), and $\bm{h}^*$ is the corresponding hint. Starting from the prefix $[\bm{y}_{<t*,},\bm{h}^*] $,  the policy model $\pi_\theta$ generates a Python snippet $\bm{c}$; the Python interpreter then returns the execution feedback $\bm{o_c}$. $\pi_\theta$ subsequently completes the partial trajectory $[\bm{y}_{<t*,}, \bm{h}^*, \bm{c},  \bm{o}_c]$ by generating $\bm{y}^{'}_{\geq t^*+\mid \bm{h}^*, \bm{c},  \bm{o_c} \mid}$. The new sub-trajectory $\bm{y}^{'}_{\geq t^*}$ is plugged  into the tree, while the original $\bm{y}$ is partitioned into $\bm{y}_{<t*}$  and $\bm{y}_{\geq t*}$, finally resulting in $\mathcal{T}_1$. In the illustrated tool‑use example, the code segment employing the \texttt{sympy} package is used to solve a function, replacing the original verbose  natural language derivation. \underline{\textbf{Upper right:}} Each leaf node, together with the nodes along the path from the root to that leaf, constitutes a complete reasoning trajectory. Tokens within a sub‑trajectory inherit the advantage value associated with that sub‑trajectory.
}
\label{fig: method}
\end{figure*}
Our approach comprises two principal sequential modules. First, we introduce rollout tree construction, a dedicated rollout mechanism devised to discover tool‑integrated reasoning trajectories (§\ref{sec:rollout_tree}). Second, we employ tree‑based advantage estimation, which leverages the constructed tree to compute the advantage associated with each sub‑trajectories (§\ref{sec:adv_estimate}) for reinforcement learning. Through these, tool-integrated reasoning behaviors can be reinforced within the policy model. 

\subsection{Rollout Tree with Tool-Integrated Forking}
\label{sec:rollout_tree}
During RL training, for a given question $\bm{q}$, we formalize the rollout as multiple reasoning with CoT to generate a group of candidate answers. Since a set of reasoning trajectories naturally forms a tree-like structure where common prefixes branch into diverse continuations, it is feasible to conceptualize reasoning trajectories generation as a tree-structured search process \cite{yang2025treerpotreerelativepolicy, li2025treepobridginggappolicy}. \textbf{Within our tree structure, each node represents a (sub-) reasoning trajectory}.

Based on this, we propose an algorithm for constructing tool-integrated rollout trees. The core idea is to iteratively expand the tree by forking new sub-reasoning trajectories that involve tool-use. As illustrated in {Fig.}~\ref{fig: method}, the rollout tree evolves through successive expansions. Forking operations are performed at positions identified as most promising within the existing reasoning trajectories---determined jointly by entropy and the likelihood of eliciting tool-use behavior---fostering exploration that integrates tool-use into the reasoning.

\noindent\textbf{Initialization}. We define the root node of the rollout tree as the question $\bm{q}$. Based on $\bm{q}$, we first generate $M$ trajectories in parallel using the policy model $\pi_{\theta}$, i.e.,
$
\bm{y} \sim \pi_\theta(\cdot \mid \bm{q})
$.
The rollout tree is initialized with the set of generated trajectories,  $\mathcal{T}_0=\{\bm{y}\}^{M}$.
The generation stops when the model generates \texttt{[EOS]} token or reaches the preset maximum length. After the initialization, the algorithm iteratively runs the following steps $N$ times for further expansion with tool-integrated trajectories. 

\noindent\textbf{Entropy-based forking position selection}. We aim to expand the tree by forking new branches (i.e., sub-trajectories) from the existing tree. We propose selecting positions with the highest uncertainty for forking. Because high-uncertainty positions enable broader exploration, whereas low-uncertainty ones typically inherit and reveal previously established lines of thought \cite{wang2025beyond}. We use entropy over vocabulary $\bm{V}$ to quantify the policy model’s uncertainty of current positions. Specifically, from the entire rollout tree, we select the top-$K$ positions $\{t\}^{K}$ that immediately follow the conclusion of a clause and have the highest entropy:
\begin{equation}
\begin{aligned}
& \{t\}^K := \underset{t\in\mathcal{T}_n}{\operatorname{arg\,top-}}K \thinspace H(t),\\
& H(t) = - \sum^{\bm{V}}_v \pi_\theta(v\mid \bm{q}, \bm{y}_{<t}) \log \pi_\theta(v\mid \bm{q}, \bm{y}_{<t}).
\end{aligned}
\end{equation}
We exclude positions near the end (within the final 20\%)  of a trajectory, since these are more associated with answer finalization than with exploration.

\noindent\textbf{Eliciting tool-integrated trajectories}. Once $\{t\}^K$
are determined, we complete the corresponding partial trajectory $\bm{y}_{<t^*}$ by resuming generation from a chosen $t^*$. Specifically, we concatenate a hint at $t^*$ to elicit a tool-integrated sub-trajectory from the policy model, yielding a novel reasoning trajectory. To this end, a set of tool-related hints $\{\bm{h}\}$ is collected, as shown in Tab.~\ref{tab:hints}, where each hint terminates with the opening delimiter of a code block to force the policy model to generate a Python code snippet. This diverse set of hints is tailored to distinct tool-use scenarios.  In order to identify a promising position-hint pair, we sample\footnote{To mitigate the tendency of the policy model to repeatedly select a limited subset of specific hints, we employ sampling instead of the argmax operation here.} a single tuple $(t^*, \bm{h}^*)$ from the joint space $ \{t\} \times \{\bm{h}\}$, according to the conditional probability $\pi_\theta(\bm{h} \mid \bm{q}, \bm{y}_{<t})$.\footnote{In the latest iteration, any $(t, \bm{h})$ selected in previous iterations are excluded. For pairs that have been probed but not selected, we cache the $\pi_\theta(\bm{h} \mid \bm{q}, \bm{y}_{<t})$ to reuse in subsequent iterations for saving computation. For this reason, a max-heap is implemented to maintain the pairs and their probabilities efficiently.} This strategy enables identifying the position $t^*$ that has substantial potential of being succeeded by the tool-use hint $\bm{h}^*$. 
With the tool-use hint $\bm{h}$ at position $t$, the policy model $\pi_\theta$ continues by generating a Python snippet $\bm{c}$ until the terminating delimiter of the code block:
\begin{equation}
\bm{c} \sim \pi_\theta(\cdot \mid \bm{q}, \bm{y}_{<t^*}, \bm{h}^*).
\label{eq:code}
\end{equation}
Upon completion, an external code sandbox with a Python interpreter receives and executes the code, and the interpreter feedback $\bm{o_c}$ is then returned. Subsequently, the policy model completes the trajectory until the model generates \texttt{[EOS]} token or reaches the preset maximum length, given as
\begin{equation}
 \bm{y}^{'}_{\geq t^*+\mid \bm{h}^*, \bm{c},  \bm{o_c} \mid}\sim \pi_\theta(\cdot \mid \bm{q}, \bm{y}_{<t^*}, \bm{h}^*, \bm{c},  \bm{o_c}).
\label{eq:continuation}
\end{equation}
$\bm{y}^{'}_{\geq t^*} = [\bm{h}^*, \bm{c},  \bm{o}_c, \bm{y}^{'}_{\geq t^*+\mid \bm{h}^*, \bm{c}, \bm{o}_c \mid}]$ is a newly generated sub-trajectory, which, together with $\bm{y}_{<t}$, forms a complete trajectory that seamlessly blends real‑time code execution and natural language. Note that, similar to the initialization, the procedure described by Eq.~\ref{eq:code} and \ref{eq:continuation} runs in parallel $M$ times for the selected $(t^*, \bm{h}^*)$, yielding the set $ \{\bm{y}^{'}_{\geq t^*}\}^M$. The rollout tree is then updated as:
\begin{equation}
\mathcal{T}_{n+1} \leftarrow \mathcal{T}_n \cup \{\bm{y}^{'}_{\geq t^*}\}^M.
\end{equation}

We perform tool-integrated forking for $N$ iteration after initializations, finally leading to the rollout tree $\mathcal{T}_N$ with $M \times (N+1)$ leaf nodes.

\subsection{Tree-based Advantage for Policy Optimization}
\label{sec:adv_estimate}
In addition to enabling the elicitation of tool‑integrated sub‑trajectories, the rollout tree affords fine-grained process supervision for intermediate steps through Monte Carlo estimation, for encouraging the beneficial tool-use behaviors that facilitate superior reasoning performance. In this section, we describe how the completed rollout tree can be leveraged to derive signals over sub-trajectories for policy optimization.

\noindent\textbf{Credit assignment for sub-trajectories}. For each leaf node $s_{leaf}$ in $\mathcal{T}$, an answer verifier takes as input the complete trajectory represented by the path from root node to $s_{leaf}$, and outputs a value $r(s_{leaf})=1$ if the answer is correct, or $0$ otherwise. 
For each non-leaf node $s_{non}$, the value is estimated via Monte Carlo methods by aggregating values from its leaf descendants. Specifically, let $Leaves(s_{non})$ denote the set of all leaf nodes that are descendants of node $s_{non}$. The value $r(s_{non})$ is calculated as the average value over its leaf descendants:
\begin{equation}
r(s_{non}) = \frac{1}{|Leaves(s_{non})|} \sum_{s \in Leaves(s_{non})}r(s)
\end{equation}
This propagates bottom-up, weighting all intermediate nodes. The value reflects the potential of the node $s$ to lead to the correct answer.

\noindent\textbf{Advantage determination}. Our goal is to derive an advantage value for each node in the rollout tree, for reflecting the relative merit of the sub-trajectories represented by that node. This advantage formulation enables reinforcement of tool‑integrated sub‑trajectories that demonstrate a comparatively higher likelihood of culminating in a correct answer, while discouraging those with lower potential. Motivated by this, we define the advantage of node $s$ as:
\begin{equation}
A(s) = \underbrace{r(s)-r(s_{root})}_{\text{Global advantage}} + \underbrace{r(s)-r(p_{s})}_{\text{Local advantage}}.
\label{eq:advantage}
\end{equation}
$s_{root}$ is the root node of the rollout tree and thus $r(s_{root})$ can be interpreted as the global accuracy of the tree. $p_s$ denotes the parent node of $s$. The global advantage of node $s$ quantifies the extent to which $s$ surpasses the tree-wide average performance. The local advantage of $s$ measures the incremental improvement $s$ brings relative to its parent. Crucially, a larger local advantage implies that $s$ is more likely than its sibling nodes to produce a correct outcome. Specifically speaking, when a new forking sub-trajectory with appropriate tool-use behavior successfully arrives at the correct answer---while its natural language siblings fail---the proposed advantage assignment strategy confers a significant advantage upon this tool-integrated sub-trajectory.

\noindent\textbf{Policy optimization}. The set of reasoning trajectories $\{\bm{S}\}^{M \times (N+1)}$ can be extracted from the rollout tree $\mathcal{T}_N$ by concatenating the sub-trajectories represented by the nodes along the path from the root to each of the $M \times (N+1)$ leaf nodes. Tokens inherit the advantage associated with the sub‑trajectory to which they belong. This process is exemplified as in the right-hand side of Fig.~\ref{fig: method}. We then adopt the on-policy Reinforce objective \cite{williams1992simple} without a KL penalty term \footnote{Other algorithms like PPO \cite{schulman2017proximal}, Reinforce++ \cite{hu2025reinforce++}, GRPO \cite{deepseek-math} can also be applicable to our setting.}, and the loss function is given as:
\begin{equation}
\begin{multlined}[0.9\columnwidth]
\mathcal{J}(\theta)
= \mathbb{E}_{q,{\{\bm{S}\}^{M \times (N+1)}}}\\
\Bigg[
\frac{1}{\sum_{i=1}^{M \times (N+1)} |\bm{S}_i|}
\sum_{i=1}^{M \times (N+1)}
\sum_{j=1}^{|\bm{S}_i|} \\
\frac{
\pi_\theta\!\left(\bm{S}_{i,j} \mid \bm{q}, \bm{S}_{i,<j}\right)
}{
\pi_{\theta_{\mathrm{old}}}\!\left(\bm{S}_{i,j} \mid \bm{q}, \bm{S}_{i,<j}\right)
}
A_{i,j}
\Bigg]
\end{multlined}
\end{equation}
$A_{i,j}$ denotes the advantage of the $j$-th token of the $i$-th trajectory. We mask out the interpreter feedback output from the loss computation to ensure training stability.

Note that tool‑integrated rollout trees are employed solely during the training phase; during testing, the inference operates in the conventional manner, since the trained model spontaneously invokes code tools.


%% file: sections/experiments.tex
\section{Experiments}
\subsection{Implementation Details}
\label{sec:setting}
\noindent\textbf{Backbone model}. The experiments are conducted with Qwen3 \cite{yang2025qwen3}, a family of natively designed large reasoning models. We employ two model variants—\texttt{Qwen3-8B} and \texttt{Qwen3-4B-Thinking-2507}—to examine the generalizability of our approach across different parameter scales. Compared to the former, the latter has fewer parameters, but has undergone more sufficient post-training and is stronger.

\noindent\textbf{Training data}. Our approach benefits from a more accurate estimation of the performance gap between natural-language and tool-integrated reasoning, and employing challenging problems can more effectively reveal improvements in tool-integrated reasoning. For these reasons, we curate a set of challenging mathematics problems for training. Specifically, we first employ \texttt{Qwen3-4B-Thinking-2507} to perform inference on all English training examples of the DAPO dataset \cite{yu2025dapo}, executing five independent runs per example. We filter out instances where the model’s accuracy exceeds 60\%, finally retaining a set of 2,795 examples as training data.

\noindent\textbf{Hyperparameters}. By default, we set $M, N, K$ to $2, 3, 6$, respectively, except in \S~\ref{sec:tree_scale}.  This setting leads to a total of 8 rollout trajectories. We employ a maximum response length of 16,384 tokens and a sampling temperature of 1.0 for both rollout and testing. See more implementation details in App.\ref{app:our_details}.

\noindent\textbf{Environment}.
Our training is conducted using VeRL \cite{Sheng_2025}, an RL training library for large language models. We implement an asynchronous code sandbox environment with SandboxFusion \cite{bytedanceseedfoundationcodeteam2025fullstackbenchevaluatingllms} \footnote{\url{https://bytedance.github.io/SandboxFusion/}}. This asynchronous setup accelerates RL training by enabling concurrent environment interactions across multiple threads, thereby maintaining high-throughput execution throughout the training.

\subsection{Baselines \& Evaluation}
We compare our approach with a comprehensive set of baseline methods. Those baselines are divided into three categories. 
1) \fbox{Direct RL} applies vanilla RL training without a cold-start stage. We use \textbf{DAPO} \cite{yu2025dapo} to train a baseline for natural language reasoning. While for TIR, we include \textbf{ToRL} \cite{li2025torl} and \textbf{UTIR} \cite{lin2025understanding}. ToRL incentivizes TIR capability in LLMs via RL, by conditioning on the TIR-related prompt and rewarding the answer correctness. UTIR additionally introduces an early-code reward to encourage early code invocation.
2) \fbox{SFT-then-RL}, before RL training, performs SFT on expert TIR trajectories to equip models with basic TIR abilities. Following \textbf{ReTool} \cite{feng2025retool}, we finetune LLMs based on expert trajectories \footnote{The expert trajectories are annotated by ReTool, available in \url{https://huggingface.co/datasets/JoeYing/ReTool-SFT}}, which consist of short CoT. After the SFT, an RL training process as ToRL is then applied. 
3) \fbox{SFT on rejection sampling}, such as \textbf{START} \cite{li-etal-2025-start}, collects tool-integrated-long CoT through rejection sampling -- using LLMs to complete long CoT after inserting tool-related hints, and retains those correct ones. The LLMs are then finetuned on the collected long CoT.  Note that it is an offline method.

For fair comparison, we use the same backbone LLMs and training data to re-produce all baselines above \footnote{For ReTool, we use the official annotated trajectories for SFT and our training data for RL.}.  In addition to \texttt{Qwen3-8B} and \texttt{Qwen3-4B-Thinking-2507}, \texttt{Qwen3-8B-Base} and \texttt{Qwen3-4B-Base} are also employed to re-produce ToRL, denoted by ToRL-base, to study the performance gaps between LRMs and LLMs (without thinking) under TIR settings. Details of the implementation of all baselines can be found in App.~\ref{app:baseline_details}.

Our in-domain evaluation benchmarks are AIME24 and AIME25, two challenging math benchmarks. GPQA-Diamond \cite{rein2023gpqagraduatelevelgoogleproofqa} is used as an out-of-domain benchmark for the evaluation on scientific questions, including physics, chemistry, and biology. Pass@$1$ and Pass@$8$ are measured for each benchmark, and the average number of code-invocations is also recorded.

\input{sections/main_table}
Two general patterns can be observed in Tab.~\ref{tab:main_experiment}. 
1) Thinking-based methods, even the ones without tool integration, are superior to tool-integrated but non-thinking counterparts, such as Text-DAPO vs. ToRL-base/ReTool, demonstrating the power of test-time scaling by long CoT \cite{openai2024openaio1card}.
2) Under thinking mode, tool-integrated approaches outperform natural language-based counterparts, such as UTIR vs. Text-DAPO, corroborating the finding by \citet{lin2025understanding} that tool-integrated-reasoning is able to expand LLMs' capability boundary.
These two findings collectively suggest that integrating tool-use into thinking trajectory (long CoTs) improves reasoning quality and leads to superior results. 

Among tool-integrated-reasoning methods, our approach shows the best performance. START relies on SFT and functions as the offline variant of our approach. Compared to our approach, it has higher code-use frequencies but suffers significant performance drops. Our manual inspection reveals that it exhibits rigid tool-use behavior that contributes little to or even misleads reasoning (see Fig.\ref{tab:tir_example_start} for an example). The degradation of its reasoning ability can be attributed to SFT on a new task with limited novel data, which greatly distorts the model’s original distribution. In comparison, \ours is trained with RL and this online training method can preserve the LLM's prior knowledge and capabilities significantly better \cite{chen2025retainingdoingroleonpolicy, shenfeld2025rlsrazoronlinereinforcement}, enabling reasonable integration of tool-use and natural language reasoning.
ToRL-thinking has virtually zero tool-use rates in its long CoT, as prompts fail to affect the underlying reasoning process.
UTIR, another RL-based method, is driven by an explicit tool-invocation reward. Our manual inspection finds  frequent void tool-use behavior (often with just a few lines of code, like  a single print function) by this method, which is mainly due to reward hacking. In our approach, the rollout chains are injected with tool-use behavior by dynamically inserting hints and are rewarded solely based on their answer correctness, effectively avoiding reward hacking. Furthermore, our hint-guided injection also fosters tool-use behavior serving a variety of objectives, thereby supporting reasoning processes (\S\ref{sec:tool_benefit}).

\section{Analysis}
We will investigate: 1) the extent to which tool-integrated reasoning benefits training; 2) the effectiveness of the tree-derived advantage; 3) the entropy-based selection of forking points; and 4) the influence brought by the scale of the rollout trees. Throughout this section, we use \texttt{Qwen3-4B-Thinking-2507} as the testbed. 
\subsection{Benefit of Tool-use for Reasoning}
\label{sec:tool_benefit}
\begin{figure}[ht]
\centering
\includegraphics[width=0.44\textwidth]{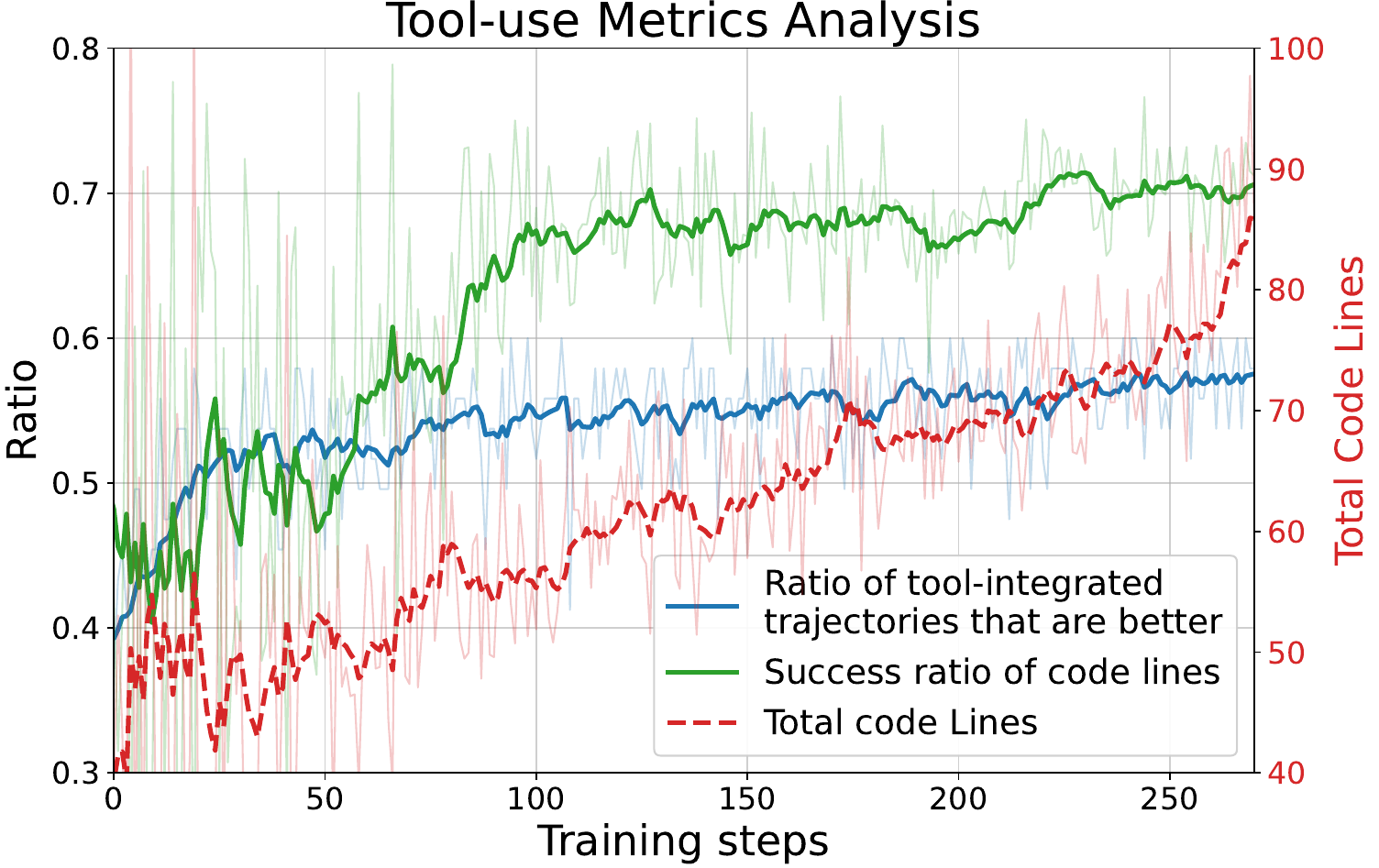}
\caption{Metrics of tool-use during training. }
\label{fig: code_analysis}
\end{figure}
A key step of our training method is to discover tool-integrated trajectories that can yield correct answers, which the original natural language-based chains are unable to achieve. The blue line in Fig.~\ref{fig: code_analysis} plots the dynamics of the proportion of training samples in which at least one tool-integrated trajectory yields the correct answer, while the natural language trajectories fail to do so. The curve starts at $0.36$ at step $0$—a non-negligible initial value—indicating that, with our hint‑based guidance, the LLM inherently possesses a capacity to employ code tools for reasoning. As training progresses, the curve rises gradually before reaching a stable plateau at approximately 0.6 (for the vast majority of the remaining 0.4, both trajectory types either succeed or fail together). Furthermore, the curves for the number of code lines (red) and the proportion of code lines that are successfully executed (green) also exhibit an upward trend. This suggests that, driven by the additional correctness afforded by tool‑integrated trajectories, the model develops a stronger tendency and enhanced capability for tool use. In turn, this promotes the discovery of tool‑integrated trajectories. App.~\ref{app:tool_use} shows analysis on tool-use by our approach during testing.

\subsection{Effectiveness of Tree-derived Process Advantage}

\begin{figure}[ht]
\centering
\includegraphics[width=0.44\textwidth]{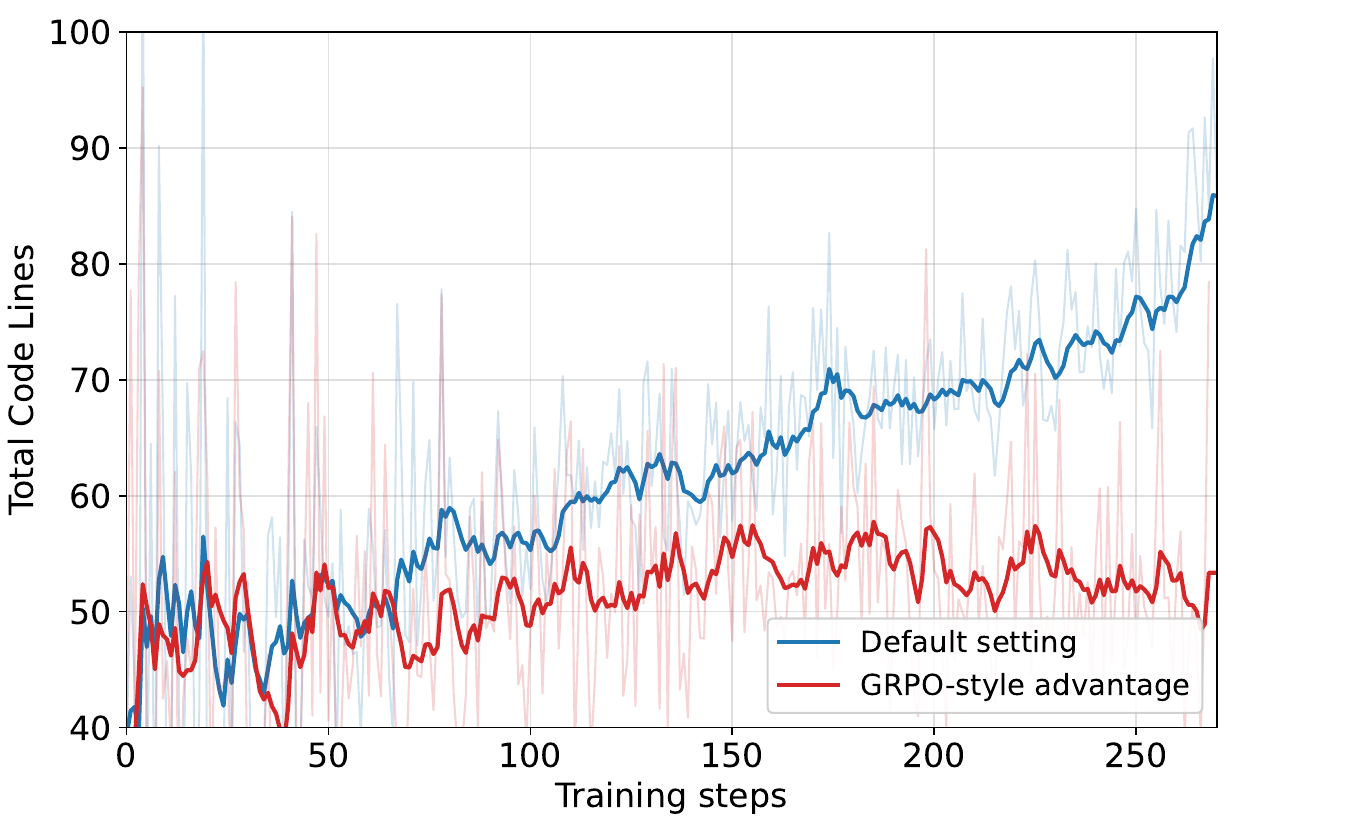}
\caption{Effect of advantage assignment on the model's tool-use tendency. }
\label{fig:adv_code_analysis}
\end{figure}

To validate the benefit of the tree-derived process advantage, we implement a variant that omits process reward. In this variant, analogous to GRPO \cite{deepseek-math}, the advantage of tokens within a trajectory is computed as the in‑group normalization of the outcome rewards among all trajectories from the rollout tree, and other settings remain the same. As shown in Block \textit{\romannumeral 2\relax} of Tab. \ref{tab:variants}, while the variant without process advantage still surpasses most baselines in \S \ref{sec:main_results}, it lags behind the default setting (tree-derived advantage) in both answer correctness and tool invocation frequency. Besides, we observe that the number of code lines of the default setting goes up more rapidly  and eventually plateaus at a higher level than that of the variant (Fig. \ref{fig:adv_code_analysis}). Based on these observations, compared with the  GRPO-style advantage assignment, where a uniform advantage is assigned to the entire sequence, our tree-derived process advantage is more effective at encouraging the model to invoke tools. This superiority arises because, under the tree‑derived advantage scheme, tokens belonging to tool‑integrated sub‑trajectories that successfully lead to correct answers receive greater advantage values than their sibling sub‑trajectories, providing discriminative signals at the forking positions.
\label{sec:grpo_adv}

\begin{table}[t]
\centering
\small
\setlength{\tabcolsep}{6pt}
\renewcommand{\arraystretch}{1.05}

\begin{tabular*}{\linewidth}{@{\extracolsep{\fill}}lccc@{}}
\toprule
& \textbf{Pass@1} & \textbf{Pass@8} & \textbf{\# Invoc.} \\
\midrule

\rowcolor{gray!15}\multicolumn{4}{l}{\textit{\textbf{i. Default setting}}} \\
\hspace{0.8em}AIME24 & 73.47 & 91.37 & 10.68 \\
\hspace{0.8em}AIME25 & 65.56 & 85.76 &  9.51 \\
\hspace{0.8em}GPQA-D & 66.65 & 83.75 &  3.81 \\
\midrule

\rowcolor{gray!15}\multicolumn{4}{l}{\textit{\textbf{ii. GRPO-style advantage}}} \\
\hspace{0.8em}AIME24 & 70.56 & 88.16 & 6.81 \\
\hspace{0.8em}AIME25 & 61.67 & 80.57 & 5.90 \\
\hspace{0.8em}GPQA-D & 64.34 & 80.94 & 3.11 \\
\midrule

\rowcolor{gray!15}\multicolumn{4}{l}{\textit{\textbf{iii. Larger rollout Tree}}} \\
\hspace{0.8em}AIME24 & 73.61 & 91.73 & 10.64 \\
\hspace{0.8em}AIME25 & 66.81 & 87.02 & 10.11 \\
\hspace{0.8em}GPQA-D & 66.24 & 83.98 &  4.17 \\
\midrule

\rowcolor{gray!15}\multicolumn{4}{l}{\textit{\textbf{iv. Random position forking}}} \\
\hspace{0.8em}AIME24 & 66.81 & 84.07 & 9.83 \\
\hspace{0.8em}AIME25 & 60.28 & 77.01 & 9.56 \\
\hspace{0.8em}GPQA-D & 61.19 & 82.95 & 3.22 \\
\bottomrule
\end{tabular*}

\caption{Comparisons across various variants of our approach. 
\textit{\textbf{\romannumeral 1\relax . Default setting}}: the default settings of our approach(\S~\ref{sec:setting});
\textit{\textbf{\romannumeral 2\relax . GRPO-style advantage}}: the advantage of tokens within a trajectory is computed as the in‑group normalization of the outcome rewards (\S~\ref{sec:grpo_adv});
\textit{\textbf{\romannumeral 3\relax . Larger rollout Tree}}: A scaled-up rollout tree with $M=3$ and $N=4$ is used for advantage estimation (\S~\ref{sec:tree_scale});
\textit{\textbf{\romannumeral 4\relax . Random position forking}}: random positions are selected for forking (\S~\ref{sec:random_selection}). For \textit{\romannumeral 2\relax}, \textit{\romannumeral 3\relax}, and \textit{\romannumeral 4\relax}, unless stated otherwise, all other settings remain the same as \textit{\romannumeral 1\relax}.
}
\label{tab:variants}
\end{table}

\subsection{Influence of Tree Scale}
\label{sec:tree_scale}
The scale of the rollout tree plays an important role in the estimation of process advantage. Given that Monte Carlo–style estimation generally benefits from larger sample sizes, we investigate whether scaling up the tree can enhance our approach’s performance by yielding a more accurate estimate of the process advantage. We set $M$ and $N$ to $3$ and $4$ respectively. For a fair comparison, we keep the experience size for policy optimization consistent by randomly sampling $8$ from a total of $15$ trajectories from the enlarged tree \footnote{The rollout trees with the default settings described in \S ~\ref{sec:setting} lead to $8$ trajectories in total}. Note that the process advantage is still estimated by using the entire tree. The results in Block \textit{\romannumeral 3\relax} of Tab. \ref{tab:variants} show that our approach with the larger rollout tree demonstrates higher pass rates across three benchmarks. 

\subsection{Effectiveness of Entropy-based Forking}
\label{sec:random_selection}
Motivated by prior findings that high‑entropy tokens steer the model toward diverse reasoning pathways and facilitate exploration \cite{wang2025beyond, treerl}, our approach selects the positions with the highest entropy as the forking points to induce tool-integrated trajectories. To assess this selection strategy, we perform an ablation by randomly choosing positions while holding other settings unchanged. The constraint in selection follows the default setting, ignoring tokens within the final 20\% of a trajectory. As evidenced by the comparison between Blocks \textit{\romannumeral 1\relax} and \textit{\romannumeral 4\relax} in Tab.~\ref{tab:variants}, this variant suffers performance degradation across three benchmarks, which substantiates the superiority of our strategy of selecting forking positions. 

%% file: sections/main_table.tex
\section{Main Results}
\label{sec:main_results}
\begin{table*}[ht]
\centering
\scalebox{0.65}{
\begin{tabular}{lllccccccccc}
\toprule \toprule
\multirow{2}{*}{\textbf{Paradigm}} & \multirow{2}{*}{\textbf{Method}} & \multirow{2}{*}{\textbf{CoT Type}} & \multicolumn{3}{c}{\textbf{AIME24}} & \multicolumn{3}{c}{\textbf{AIME25}} & \multicolumn{3}{c}{\textbf{GPQA-D}} \\
\cmidrule(lr){4-6} \cmidrule(lr){7-9} \cmidrule(lr){10-12}
 & & & \textbf{Pass@1} & \textbf{Pass@8} & \textbf{\#Invoc.} & \textbf{Pass@1} & \textbf{Pass@8} & \textbf{\#Invoc.} & \textbf{Pass@1} & \textbf{Pass@8} & \textbf{\#Invoc.} \\
\midrule
\multicolumn{12}{l}{{\large\textit{\textbf{Qwen3-4B-Thinking-2507}}}} \\
 & Original checkpoint & pure-text long & 52.22 & 71.93 & 0 & 46.39 & 65.00 & 0 & 65.98 & 83.32 & 0 \\
\midrule
\multirow{4}{*}{RL} & DAPO & pure-text long & 69.17 & 85.67 & 0 & 61.53 & 84.09 & 0 & 65.53 & 82.82 & 0 \\
 & ToRL-Think & pure-text long & 71.25 & 82.14 & 0 & 61.25 & 76.27 & 0 & 65.15 & 74.99 & 0 \\
 & ToRL-Base & tool-integrated short & 40.49 & 50.73 & 3.12 & 31.81 & 42.24 & 2.99 & 31.67 & 43.65 & 1.01 \\
 & UTIR & tool-integrated long & 70.56 & 87.61 & 10.08 & 63.89 & 77.96 & 8.12 & 66.08 & 81.90 & 2.11 \\
\midrule
\multirow{2}{*}{SFT} & Retool (only SFT) & tool-integrated short & 17.36 & 34.14 & 2.12 & 13.47 & 23.42 & 1.89 & 20.12 & 42.06 & 1.12 \\
 & START & tool-integrated long & 44.31 & 65.15 & 13.30 & 38.47 & 58.90& 12.08 & 55.82 & 82.22 & 4.24 \\
\midrule
SFT+RL & Retool & tool-integrated short & 23.33 & 35.46 & 1.85 & 20.13 & 36.36 & 1.35 & 13.26 & 20.95 & 0.98 \\
\midrule
\rowcolor{gray!15}
RL & \textbf{Ours} & tool-integrated long & \textbf{73.47} & \textbf{91.37} & 10.68 & \textbf{65.56 }& \textbf{85.76 }& 9.51 & \textbf{66.65 } & \textbf{83.75 } & 3.81 \\
\midrule
\midrule
\multicolumn{12}{l}{{\large\textit{\textbf{Qwen3-8B}}}} \\
 & Original checkpoint & pure-text long & 64.17 & 79.28 & 0 & 52.08 & 69.22 & 0 & 58.88 & 79.53 & 0 \\
\midrule
\multirow{4}{*}{RL} & DAPO & pure-text long & 66.11 & 81.44 & 0 & 56.11 & 76.98 & 0 & 59.41 & 80.29 & 0 \\
 & ToRL-Think & pure-text long & 65.70 & 82.68 & 0 & 59.44 & 69.91 & 0 & 55.24 & 67.11 & 0 \\
 & ToRL-Base & tool-integrated short & 42.64 & 54.95 & 3.86 & 25.83 & 37.28 & 2.98 & 41.56 & 56.62 & 1.06 \\
 & UTIR & tool-integrated long & \textbf{69.12} & 84.59 & 11.97 & \textbf{61.11 } & 74.96 & 10.22 & 59.70 & 80.71 & 2.74 \\
\midrule
\multirow{2}{*}{SFT} & Retool (only SFT) & tool-integrated short & 22.78 & 43.46 & 2.00 & 19.67 & 31.18 & 2.07 & 21.23 & 46.32 & 1.16 \\
 & START & tool-integrated long & 35.83 & 58.22 & \textbf{15.08 }& 30.28 & 49.76 & 12.77 & 35.89 & 67.61 & 5.79 \\
\midrule
SFT+RL & Retool & tool-integrated short & 25.41 & 46.73 & 1.67 & 30.00 & 43.51 & 1.20 & 18.27 & 36.04 & 0.92 \\
\midrule
\rowcolor{gray!15}
RL & \textbf{Ours} & tool-integrated long & 68.89 & \textbf{86.81} & 12.82 & 57.64 & \textbf{79.02 } & 10.66 & \textbf{60.03  }& \textbf{81.57} & 4.14 \\
\bottomrule \bottomrule
\end{tabular}
}
\caption{Comparison of different methods on AIME24, AIME25, and GPQA-Diamond benchmarks. \textit{\# Invoc.} denotes the average number of code-invocations in short/long CoT on the test set. For Retool and ToRL-Base, which rely on short CoT, the value is measured in short CoT; for other methods, it is measured in long CoT. For ToRL-Think, tool-invocations occur only outside the thinking part of the responses.}
\label{tab:main_experiment}
\end{table*}

%% file: sections/related_works.tex
\section{Related Works}
Tool-Integrated Reasoning (TIR) has emerged as a pivotal paradigm to address the inherent limitations of Large Language Models (LLMs) in precise computation, symbolic manipulation, and complex algorithm execution, by augmenting LLMs with external tools such as code interpreters, search engines, and numerical libraries. Early advancements in TIR primarily relied on Supervised Fine-Tuning (SFT) to imprint tool-use patterns into models: \citet{yang2024qwen25mathtechnicalreportmathematical} synthesized data with tool-integrated responses for SFT, enabling Python interpreter integration and achieving superior performance on mathematical benchmarks; ToRA \cite{gou2024toratoolintegratedreasoningagent} and MathCoder \cite{wang2023mathcoderseamlesscodeintegration} demonstrated the efficacy of code execution in enhancing mathematical reasoning accuracy, laying the foundation for tool-augmented symbolic computation. These SFT-based approaches, however, are constrained by predefined tool-use trajectories, limiting generalization to unseen tasks or adaptive strategy discovery.
START \cite{li-etal-2025-start} further innovated TIR with self-training pipelines: it activates latent tool-use capabilities by synthesizing high-quality training data through diverse hint patterns and rejection sampling.

Recent research has shifted toward Reinforcement Learning (RL)-driven TIR, enabling models to autonomously explore optimal tool-use strategies via outcome feedback. ReTool \cite{feng2025retool} proposed a two-stage framework: cold-start SFT with code-augmented reasoning traces, followed by RL with interleaved real-time code execution. ToRL \cite{li2025torl} further scaled RL directly from base models without prior SFT, unlocking emergent behaviors of tool invocation for computational and analytical reasoning. Complementary to these, AutoTIR \cite{wei2025autotir} designed a hybrid reward mechanism to balance tool integration with core language capabilities, supporting autonomous selection among multiple tools (e.g., search engines, code interpreters) and achieving superior generalization across knowledge-intensive, mathematical, and instruction-following tasks.
\citet{lin2025understanding} provided the first formal theoretical proof that TIR strictly expands an LLM’s empirical and feasible support; they also proposed Advantage Shaping Policy Optimization to stably guide tool-use behavior (e.g., early code invocation).

Beyond mathematical reasoning, TIR has been extended to various specialized scenarios: Search-R1 \cite{jin2025searchr} and WebSailor \cite{tongyidr} integrated retrieval tools with RL for multi-hop QA and web agent tasks, respectively.  \citet{chai2025scimastergeneralpurposescientificai} orchestrated domain-specific tools for tasks such as automated  scientific data analysis.

%% file: sections/appendix.tex
\section{Implementation Details of Our Approach}
\label{app:our_details}
\subsection{Prompt Template}
\label{app:our_prompt}
The prompt template for our approach is shown below. \{\texttt{Question}\} will be replaced with specific questions during training and inference.
\begin{prompt}[notitle]{-15pt}{-5pt}{}
    <|im_start|>user
    Solve the following problem. You can use \boxed to return your answer. The last part of your response should be:
    \boxed{'The final answer goes here.'}
    
    {Question}
    <|im_end|>
    <|im_start|>assistant
\end{prompt}

\subsection{Hints}
We use a collection of hints for code‑use guidance compiled by \cite{li-etal-2025-start}. Each hint targets a distinct application scenario, such as complex calculations or self‑reflection. During rollout, each hint is probed at positions with the highest entropy to identify a promising position–hint pair, whereupon the hint is inserted at the selected position (\S \ref{sec:rollout_tree}). 

\begin{table}[h]
\centering
\scalebox{0.75}{
\begin{tabular}{p{9.5cm}}
\toprule
\textbf{Complex calculations} \\
\texttt{I can use Python to perform complex calculations for this problem.\textbackslash n\textasciigrave\textasciigrave\textasciigrave python}
\\\hline
\textbf{Self-reflection} \\
\texttt{I can use Python to check if my approach is correct and refine it, if necessary.\textbackslash n\textasciigrave\textasciigrave\textasciigrave python}
\\\hline
\textbf{Check logic}  \\
\texttt{maybe Python can assist in ensuring our logical deductions are sound.\textbackslash n\textasciigrave\textasciigrave\textasciigrave python}
\\\hline
\textbf{Alternative method}  \\
\texttt{I can use Python to explore an alternative method for solving this problem.\textbackslash n\textasciigrave\textasciigrave\textasciigrave python}
\\\hline
\textbf{General}  \\
\texttt{maybe using python here is a good idea.\textbackslash n\textasciigrave\textasciigrave\textasciigrave python}
\\\hline
\textbf{Deeper think}  \\
\texttt{I can think more deeply about this problem through python tools.\textbackslash n\textasciigrave\textasciigrave\textasciigrave python}
\\ \bottomrule
\end{tabular}}
\caption{The collection of hints, curated by \citet{li-etal-2025-start}.}
\label{tab:hints}
\end{table}

\subsection{Training Hyperparameters}
\label{app:hyperpara}
Tab.\ref{tab:hyperpara} lists the hyperparameters for RL training. The rollout runs in the SGLang \cite{zheng2024sglangefficientexecutionstructured} inference engine under asynchronous settings to speed up the training.

\begin{table}[h]
\centering
\scalebox{0.65}{
\begin{tabular}{ll}
\toprule
\textbf{Hyperparameters}  \\ \hline
Training batch size & 32 \\
Optimizer & AdamW \\
 &  \cite{Loshchilov2017DecoupledWD}\\
Learning rate & 1e-7 \\
Warmup step & 0 \\
Gradient accumulation step & 1 \\
Learning rate scheduler & Linear \\
KL coefficient & 0 \\
Rollout temperature & 1 \\ 
Rollout maximum length & 16,384 \\
Rollout number & 8 by default, depending on rollout trees \\ 
Total epoch & 5 \\ \bottomrule
\end{tabular}}
\caption{The hyperparameters used in our RL training.}
\label{tab:hyperpara}
\end{table}

\subsection{Interaction with Environment}
We employ SandboxFusion to build a sandbox integrated with a Python interpreter. This environment executes Python code generated by the model during the reasoning process and returns the corresponding results. To manage output length, any overlong result is truncated to a maximum of 512 tokens, preserving the final segment.

\section{Implementation Details of Baselines}
\label{app:baseline_details}
\subsection{Text-DAPO}
Text-DAPO uses the same prompt and training hyperparameters as our approach (\ref{app:our_details}). In addition, it adopts dynamic sampling to enhance training stability: prompt groups whose accuracy is exactly 1 or 0 are filtered out. The overlong penalty factor is set to 1.0.

\subsection{ToRL}
\label{app:torl_details}
We re-use the prompt template in the official code, as shown below. All other settings for RL strictly follow those of our approach (App.\ref{app:hyperpara}).
\begin{prompt}[notitle]{-15pt}{-5pt}{}
    <|im_start|>system
    A conversation between User and Assistant. The user asks a question, and the Assistant solves it.
    <|im_end|>
    <|im_start|>user
    Please integrate natural language reasoning with programs to solve the problem above, and put your final
    answer within \boxed{}.
    
    {Question}

    <|im_end|>
    <|im_start|>assistant
\end{prompt}

\subsection{ReTool}
ReTool consists of sequential stages of supervised fine‑tuning followed by reinforcement learning. For SFT, we finetune the models with the annotated tool-integrated short CoT, provided by \citet{feng2025retool}. The training set consists of 2,000 samples on math problems. We employ LlamaFactory \cite{zheng2024llamafactory} to perform SFT, with the training hyperparameters shown in Tab.~\ref{tab:retool_hyperpara}.

\begin{table}[h]
\centering
\scalebox{0.75}{
\begin{tabular}{ll}
\toprule
\textbf{Hyperparameters}  \\ \hline
Training batch size & 16 \\
Optimizer & AdamW \\
Learning rate & 1e-4 \\
Warmup ratio & 0.1 \\
Gradient accumulation step & 3 \\
Learning rate scheduler & Cosine \\
Maximum length & 16,384 \\ 
Total epoch & 3 \\ \bottomrule
\end{tabular}}
\caption{The hyperparameters used in the SFT stage of ReTool and START.}
\label{tab:retool_hyperpara}
\end{table}

In order to retain the competencies learned in SFT, the same prompt template from the SFT data is employed during reinforcement learning training, as illustrated below. All other settings for RL strictly follow those of our approach (App.\ref{app:hyperpara}).

\begin{prompt}[notitle]{-15pt}{-5pt}{}
    <|im_start|>user
    Solve the following problem step by step. You now have the ability to selectively write executable Python code to enhance your reasoning process. The Python code will be executed by an external sandbox, and the output (wrapped in `<interpreter>output_str</interpreter>`) can be returned to aid your reasoning and help you arrive at the final answer. The Python code should be complete scripts, including necessary imports. \nEach code snippet is wrapped with `<code>\n```python\ncode snippet\n```\n</code>`.\nThe last part of your response should be in the following format:\n<answer>\n\\boxed{{'The final answer goes here.'}}\n</answer>\n\n*user question:*
    
    {Question}

    Remember to place the final answer in the last part using the format: \n<answer>\n\\boxed{{'The final answer goes here.'}}
    <|im_end|>
    <|im_start|>assistant
\end{prompt}

\subsection{START}

The original work has not been open‑sourced yet. To reproduce START \cite{li-etal-2025-start}, we follow the default settings reported in the original paper, but uses our own backbone models and training data (\S~\ref{sec:setting}) to ensure a fair comparison. For collecting self‑training data with tool‑integrated long CoT, we use \texttt{Qwen3-8B} and \texttt{Qwen3-4B-Thinking-2507}\footnote{QwQ-32B-Preview was used in the original paper.} to perform inference ten times for each question in the training set. Guided by the hints in Tab.\ref{tab:hints}, the models generate reasoning-aiding code. We retain only those generated trajectories that both yield correct answers and involve tool use, ultimately obtaining 11,360 samples. Self‑training on these data is conducted with the hyperparameter set listed in Tab.~\ref{tab:retool_hyperpara}.

\subsection{UTIR}
We use the official prompt template as shown below. 
\begin{prompt}[notitle]{-15pt}{-5pt}{}
    <|im_start|>system
    Solve the following problem step by step. You now have the ability to selectively write executable Python code to enhance your reasoning process. The Python code will be executed by an external sandbox, and the output (wrapped in ```output_str```) can be returned to aid your reasoning and help you arrive at the final answer. The Python code should be complete scripts, including necessary imports. Important: The sandbox is stateless and non-interactive; thus, prior imports, definitions, and state do not persist between executions and cannot be referenced.\nEach code snippet is wrapped with ```python\ncode snippet\n```.\nThe last part of your response should be in the following format:
    
    <answer>
    \\boxed{{'The final answer goes here.'}}
    </answer>

    <|im_end|>
    <|im_start|>user
    {Question}

    Remember to place the final answer in the last part using the format: 
    <answer>
    \\boxed{{'The final answer goes here.'}}
    </answer>
    <|im_end|>
    <|im_start|>assistant

\end{prompt}

For early-code reward, we set $\delta$ to -2.0 and $k$ to 0.7 for a conservative setting. We refer readers to \citet{lin2025understanding} for further details of this method. All other settings for RL training follow those of our approach (App.\ref{app:hyperpara}).

\section{Tool-use Analysis}
\label{app:tool_use}
\subsection{Python Packages}
Tab.~\ref{tab:python_math_libs} lists the Python libraries most frequently used by our models during AIME testing, which are primarily specialized in mathematical and scientific computing.

\begin{table}[h]
\centering
\scalebox{0.75}{
\begin{tabular}{lp{6.5cm}}
        \toprule
        \textbf{Package} & \textbf{Description} \\
        \midrule
        \texttt{math} & standard mathematical functions for  arithmetic \\
        \texttt{numpy} & calculation of multi-dimensional arrays and matrices \\
        \texttt{complex}  & mathematical functions specifically for complex numbers  \\
        \texttt{networkx} & creation, manipulation, and study of complex networks and graph structures \\
        \texttt{pulp} &  generation of linear programming problems and calling of external solvers \\
        \texttt{gmpy2} & fast, multiple-precision arithmetic \\
        \texttt{pandas} & data structures and data analysis tools \\
        \texttt{divisors} & specialized utility for efficiently generating divisors of integers \\
        \texttt{primefactors} &   prime factorization of numbers \\
        \texttt{complex-number} & complex number arithmetic  \\
        \texttt{symbolic} & symbolic mathematics and algebraic manipulation \\
        \texttt{sympy} & symbolic mathematics and computer algebra \\
        \texttt{scipy} & scientific computing, including optimization, integration, and statistics \\
        \texttt{shapely} &  manipulation and analysis of planar geometric objects \\
        \bottomrule
\end{tabular}}
\caption{Python libraries for mathematical and scientific computing used by our models during testing on AIME.}
\label{tab:python_math_libs}
\end{table}

\subsection{Case Study}
We provide cases answered by \textsc{DART} (Tab.~\ref{tab:tir_example_dart1}, \ref{tab:tir_example_dart2}, \ref{tab:tir_example_dart3}) and START (Tab.~\ref{tab:tir_example_start}) respectively.
\input{sections/case_study}
\input{sections/algorithm}

%% file: sections/case_study.tex
\newpage
\onecolumn
{\footnotesize
\begin{llmcontainer}
    \noindent\textbf{[Question]}
    The $9$ members of a baseball team went to an ice-cream parlor after their game. Each player had a single scoop cone of chocolate, vanilla, or strawberry ice cream. At least one player chose each flavor, and the number of players who chose chocolate was greater than the number of players who chose vanilla, which was greater than the number of players who chose strawberry. Let $N$ be the number of different assignments of flavors to players that meet these conditions. Find the remainder when $N$ is divided by $1000.$

    \vspace{2pt}
    \noindent\rule{\linewidth}{0.5pt}
    
    \noindent\textbf{[Response]}\\
    <think>\\
    \dots
    
    Now, 2016 divided by 1000 is 2 with a remainder of 16. So the remainder is 16.

    Wait, but I just have a tiny doubt because sometimes problems like this can have tricks, like if they said at least one player chose each flavor but maybe I misinterpreted that as s $\geq$ 1, which I did, and the inequalities are strict, which I did. Let me check an external source in my mind: for example, if it was non-strict, but no, problem says greater than, so strict inequalities.
    \vspace{1em}
    Is there any chance the problem meant ordered by flavor counts but we need to consider permutations? But no—we fixed c as chocolate, v as vanilla, s as strawberry, so each flavor has a specific count, so we don't have to multiply by permutations for flavor labels because the flavors are distinct and labeled (chocolate, vanilla, strawberry are different flavors). For example, if we were counting for unlabeled flavors, it would be different, but here each flavor is distinct: chocolate is a specific flavor, so assigning 6 to chocolate is a different count than assigning 6 to vanilla, but in our problem, we're just counting assignments where the numbers satisfy c > v > s for chocolate > vanilla > strawberry, so we don't have to permute the counts among flavors—each flavor is fixed in its role (chocolate is the one with most, vanilla medium, strawberry least).
    \vspace{1em}
    Another check: suppose we had a smaller case to see if our method is correct. But maybe too time-consuming. Alternatively, let's think: is 2016 the right total?
    \vspace{1em}
    Wait, maybe I can use Python to verify the calculations:

    \textasciigrave\textasciigrave\textasciigrave \texttt{python}
    
    \begin{lstlisting}[language=Python, label={lst:polygon_solver}, basicstyle=\ttfamily\scriptsize]
# Calculate multinomial coefficients for each valid (c, v, s) triple
from math import factorial

# Case 1: (6, 2, 1)
case1 = factorial(9) // (factorial(6) * factorial(2) * factorial(1))

# Case 2: (5, 3, 1)
case2 = factorial(9) // (factorial(5) * factorial(3) * factorial(1))

# Case 3: (4, 3, 2)
case3 = factorial(9) // (factorial(4) * factorial(3) * factorial(2))

N = case1 + case2 + case3
remainder = N % 1000

print(f""Case 1: {case1}, Case 2: {case2}, Case 3: {case3}, N: {N}, Remainder: {remainder}"")
    \end{lstlisting}
\textasciiacute \textasciiacute \textasciiacute\\
\vspace{1em}
I receive the execution result from python interpreter: \\
    \textasciigrave\textasciigrave\textasciigrave
    \begin{lstlisting}[language=Python, label={lst:polygon_solver}, basicstyle=\ttfamily\scriptsize]
Case 1: 252, Case 2: 504, Case 3: 1260, N: 2016, Remainder: 16
    \end{lstlisting}
\textasciiacute \textasciiacute \textasciiacute\\
\vspace{1em}

Yes, Python confirms it. So my initial calculation is correct. The remainder when 2016 is divided by 1000 is 16. I think I was overworried because sometimes problems can be tricky, but in this case, all steps check out—found all valid distributions using integer partitions with strict inequalities, verified each multinomial coefficient with multiple methods, cross-checked with Python (which is a good way to avoid arithmetic errors), and the total is definitely 2016. Dividing by 1000 gives a remainder of 16.

So the answer should be 16.\\
</think>\\
...
\end{llmcontainer}
}
\captionof{table}{A case by \textsc{DART} (Qwen3-4B-Thinking): in the long CoT, a python snippet involving \textsc{math.factorial} is used to perform calculation for self-verification.}\label{tab:tir_example_dart1}

\newpage
{\footnotesize
\begin{llmcontainer}
    \noindent\textbf{[Question]}
The set of points in $3$-dimensional coordinate space that lie in the plane $x+y+z=75$ whose coordinates satisfy the inequalities\[x-yz<y-zx<z-xy\]forms three disjoint convex regions. Exactly one of those regions has finite area. The area of this finite region can be expressed in the form $a\sqrt{b},$ where $a$ and $b$ are positive integers and $b$ is not divisible by the square of any prime. Find $a+b.$

    \vspace{2pt}
    \noindent\rule{\linewidth}{0.5pt}
    
    \noindent\textbf{[Response]}\\
    <think>\\
    \dots
    
    Wait, the problem says exactly one region has finite area. In a plane with four lines, the maximum number of bounded regions is 4 (if they form a convex quadrilateral with diagonals), but here we have six intersection points, so it's a convex polygon with six vertices? Wait, no—four lines in general position form a quadrilateral (4 sides, 4 vertices), but here four lines: L1, L2, L3, L4—do they form a convex quadrilateral or more?
    \vspace{1em}
    Let's use the Python shell to check for intersections.

    \textasciigrave\textasciigrave\textasciigrave \texttt{python}
    
    \begin{lstlisting}[language=Python, label={lst:polygon_solver}, basicstyle=\ttfamily\scriptsize]
from sympy import symbols, Eq, solve

x, y = symbols('x y')

# Define all four boundary lines (in xy-plane)
L1 = Eq(x, y)  # x = y
L2 = Eq(x + y, 76)  # x + y = 76
L3 = Eq(x + 2*y, 75)  # x + 2y = 75
L4 = Eq(x, -1)  # x = -1

# Find all intersections (6 points)
intersections = []
intersections.append(solve((L1, L2), (x, y)))
intersections.append(solve((L1, L3), (x, y)))
intersections.append(solve((L1, L4), (x, y)))
intersections.append(solve((L2, L3), (x, y)))
intersections.append(solve((L2, L4), (x, y)))
intersections.append(solve((L3, L4), (x, y)))

print(intersections)
    \end{lstlisting}
\textasciiacute \textasciiacute \textasciiacute\\
\vspace{1em}
I receive the execution result from python interpreter: \\
    \textasciigrave\textasciigrave\textasciigrave
    \begin{lstlisting}[language=Python, label={lst:polygon_solver}, basicstyle=\ttfamily\scriptsize]
[{x: 38, y: 38}, {x: 25, y: 25}, {x: -1, y: -1}, {x: 77, y: -1}, {x: -1, y: 77}, {x: -1, y: 38}]
    \end{lstlisting}
\textasciiacute \textasciiacute \textasciiacute\\
\vspace{1em}
This gives the six intersection points in the xy-plane:\\
1. (38, 38)\\
2. (25, 25)\\
3. (-1, -1)\\
4. (77, -1)\\
5. (-1, 77)\\
6. (-1, 38)\\
...\\
</think>\\
...
\end{llmcontainer}
}
\captionof{table}{A case by \textsc{DART} (Qwen3-4B-Thinking): in the long CoT, a Python snippet utilizes \textsc{SymPy} to symbolically solve for the intersection points of boundary lines.}\label{tab:tir_example_dart2}

\newpage
\onecolumn
{\footnotesize
\begin{llmcontainer}
    \noindent\textbf{[Question]}
There are $8!= 40320$ eight-digit positive integers that use each of the digits $1, 2, 3, 4, 5, 6, 7, 8$ exactly once. Let $N$ be the number of these integers that are divisible by $22$. Find the difference between $N$ and $2025$.

    \vspace{2pt}
    \noindent\rule{\linewidth}{0.5pt}
    
    \noindent\textbf{[Response]}\\
    <think>\\
    \dots\\
    I can use Python to define a function that counts the number of even digits in each subset.\\
    \textasciigrave\textasciigrave\textasciigrave \texttt{python}
    
    \begin{lstlisting}[language=Python, label={lst:polygon_solver}, basicstyle=\ttfamily\scriptsize]
def count_evens_in_subsets(subsets):
    evens = [2, 4, 6, 8]
    total_evens = 0
    for subset in subsets:
        # Subset is a tuple, convert to list for easy access
        subset_list = list(subset)
        count = sum(1 for digit in subset_list if digit in evens)
        total_evens += count
    return total_evens

total_evens = count_evens_in_subsets(subsets)
print(total_evens)
    \end{lstlisting}
\textasciiacute \textasciiacute \textasciiacute\\
\vspace{1em}
I get the error message from python interpreter:  \\
    \textasciigrave\textasciigrave\textasciigrave
    \begin{lstlisting}[language=Python, label={lst:polygon_solver}, basicstyle=\ttfamily\scriptsize]
NameError: name 'subsets' is not defined
    \end{lstlisting}
\textasciiacute \textasciiacute \textasciiacute\\
\vspace{1em}
Let me correct that.

    \textasciigrave\textasciigrave\textasciigrave \texttt{python}
    
    \begin{lstlisting}[language=Python, label={lst:polygon_solver}, basicstyle=\ttfamily\scriptsize]
from itertools import combinations

def find_subsets():
    digits = list(range(1, 9))
    subsets = []
    for subset in combinations(digits, 4):
        if sum(subset) == 18:
            subsets.append(subset)
    return subsets

subsets = find_subsets()
evens = [2, 4, 6, 8]

def count_evens_in_subsets(subsets, evens):
    total_evens = 0
    for subset in subsets:
        subset_list = list(subset)
        count = sum(1 for digit in subset_list if digit in evens)
        total_evens += count
    return total_evens

total_evens = count_evens_in_subsets(subsets, evens)
print(total_evens)
    \end{lstlisting}
\textasciiacute \textasciiacute \textasciiacute\\
\vspace{1em}
I receive the execution result from python interpreter:   \\
    \textasciigrave\textasciigrave\textasciigrave
    \begin{lstlisting}[language=Python, label={lst:polygon_solver}, basicstyle=\ttfamily\scriptsize]
16
    \end{lstlisting}
\textasciiacute \textasciiacute \textasciiacute\\
\vspace{1em}
Perfect, so the code confirms it: each of the 8 subsets has 2 even digits, so total count of even digits across all subsets is 16.
...\\
</think>\\
...
\end{llmcontainer}
}
\captionof{table}{A case by \textsc{DART} (Qwen3-4B-Thinking): the code initially generated by DART contained a bug. After receiving feedback from the interpreter, it was tried again and then written correctly}\label{tab:tir_example_dart3}

\newpage
{\footnotesize
\begin{llmcontainer}
    \noindent\textbf{[Question]}
Every morning Aya goes for a $9$-kilometer-long walk and stops at a coffee shop afterwards. When she walks at a constant speed of $s$ kilometers per hour, the walk takes her 4 hours, including $t$ minutes spent in the coffee shop. When she walks $s+2$ kilometers per hour, the walk takes her 2 hours and 24 minutes, including $t$ minutes spent in the coffee shop. Suppose Aya walks at $s+\frac{1}{2}$ kilometers per hour. Find the number of minutes the walk takes her, including the $t$ minutes spent in the coffee shop.

    \vspace{2pt}
    \noindent\rule{\linewidth}{0.5pt}
    
    \noindent\textbf{[Response]}\\
    <think>\\
    \dots
    
    Now, I can use Python to check if my approach is correct.

    \textasciigrave\textasciigrave\textasciigrave \texttt{python}
    
    \begin{lstlisting}[language=Python, label={lst:polygon_solver}, basicstyle=\ttfamily\scriptsize]
# Let's set up the equations in minutes to be thorough

# First scenario: v = s, total time T1 = 240 minutes
# Second scenario: v = s + 2, total time T2 = 144 minutes

# So:
# 540/s + t = 240  --> equation 1
# 540/(s + 2) + t = 144 --> equation 2

# Subtract equation 2 from equation 1: 540/s - 540/(s + 2) = 240 - 144 = 96
    \end{lstlisting}
\textasciiacute \textasciiacute \textasciiacute\\
\vspace{1em}
This code does not return any output, perhaps because there is no print statement.\\
...\\
</think>\\
...
\end{llmcontainer}
}
\captionof{table}{A case by \textsc{START} (Qwen3-4B-Thinking): in the long CoT, the generated Python code did not return any valid information.}\label{tab:tir_example_start}

%% file: sections/algorithm.tex
\begin{algorithm*}[hbt!]
\caption{DART: Rollout Tree Construction and Advantage Estimation}
\label{alg:dart_single}
\begin{algorithmic}[1]
\Require 
    Question $q$; 
    Policy model $\pi_\theta$; 
    Set of tool-use hints $\{\bm{h}\}$; 
    Tree width $M$ (chains per fork); 
    Max expansion depth $N$; 
    Top-$K$ entropy positions.

\Ensure Set of trajectories with advantages $\{(S, A(S))\}$

\State \textbf{// Phase 1: Rollout Tree Construction}
\State Initialize tree $\mathcal{T}_0$ with root node $q$
\State Generate $M$ initial reasoning chains: $\mathcal{Y}_0 = \{\mathbf{y}_1, \dots, \mathbf{y}_M\} \sim \pi_\theta(\cdot | q)$
\State $\mathcal{T}_0 \leftarrow \mathcal{T}_0 \cup \mathcal{Y}_0$

\For{$n = 1$ to $N$}
    \State \textbf{Step 1.1: Entropy-based Forking Position Selection}
    \State Select candidate positions $\{t\}^K$ with highest entropy
    
    \State \textbf{Step 1.2: Hint Selection \& Branching}
    \State Sample position-hint $\{t^*,h^*\} \sim \pi_\theta(h | q, \mathbf{y}_{<t^*})$ from $\{\bm{h}\}$
    \State Generate code snippet: $c \sim \pi_\theta(\cdot \mid q, \mathbf{y}_{<t^*}, h^*)$ (Eq. \ref{eq:code})
    \State Execute code in sandbox: $o_c \leftarrow \text{PythonInterpreter}(c)$
    \State Generate continuation: $\mathbf{y}' \sim \pi_\theta(\cdot \mid q, \mathbf{y}_{<t^*}, h^*, c, o_c)$ (Eq. \ref{eq:continuation})
    \State Form new sub-trajectory: $\mathbf{y}_{new} = [h^*, c, o_c, \mathbf{y}']$
    
    \State Update tree: $\mathcal{T}_n \leftarrow \mathcal{T}_{n-1} \cup \{\text{all } \mathbf{y}_{new}\}$
\EndFor

\State Let $\mathcal{T}_{final} = \mathcal{T}_N$ be the completed rollout tree

\State \textbf{// Phase 2: Tree-based Advantage Estimation}
\State \textbf{Step 2.1: Leaf Evaluation}
\For{each leaf node $s_{leaf} \in \mathcal{T}_{final}$}
    \State Verify answer correctness: $r(s_{leaf}) \leftarrow \mathbb{I}(\text{answer is correct})$
\EndFor

\State \textbf{Step 2.2: Value Propagation (Bottom-up)}
\For{each non-leaf node $s$ in $\mathcal{T}_{final}$ (from leaves to root)}
    \State $r(s) \leftarrow \frac{1}{|\text{Leaves}(s)|} \sum_{l \in \text{Leaves}(s)} r(l)$ \Comment{Avg correctness of descendants }
\EndFor

\State \textbf{Step 2.3: Advantage Calculation}
\State Extract all nodes/sub-chains $\{S\}$ from $\mathcal{T}_{final}$
\For{each node $s \in \{S\}$}
    \State Global Advantage: $A_{global} = r(s) - r(s_{root})$
    \State Local Advantage: $A_{local} = r(s) - r(\text{parent}(s))$
    \State Total Advantage: $A(s) \leftarrow A_{global} + A_{local}$ \Comment{Reflects both global and incremental merit}
\EndFor

\State \Return Trajectories and token-level advantages $\{(S, A(S))\}$ for optimization
\end{algorithmic}
\end{algorithm*}